\documentclass[runningheads]{llncs}

\usepackage{eccv}

\usepackage{eccvabbrv}

\usepackage{graphicx}
\usepackage{booktabs}

\usepackage[accsupp]{axessibility}  

\usepackage{multirow}
\usepackage{rotating}   
\usepackage{array}
\usepackage{tabularx}
\usepackage[table]{xcolor} 

\usepackage{hyperref}

\usepackage{orcidlink}

\definecolor{bestconf}{rgb}{0.12, 0.49, 0.85} 
\definecolor{secondconf}{RGB}{255, 210, 166} 
\newcommand{\best}[1]{\cellcolor{bestconf!30}\textbf{#1}}
\newcommand{\second}[1]{\cellcolor{bestconf!12}\underline{#1}}

\newcolumntype{Y}{>{\centering\arraybackslash}X}

\begin{document}

\title{Geometry-Preserving in 3D Gaussian Splatting for LiDAR-Camera Extrinsic Calibration
} 

\titlerunning{Geometry-Preserving 3DGS for LiDAR-Camera Calibration}

\author{Kyoleen Kwak\inst{1} \and
Daeho Kim\inst{1} \and
Jeong Woon Lee\inst{1} \and
Hyoseok Hwang\thanks{Corresponding author.}\inst{1} }

\authorrunning{K. Kwak et al.}

\institute{Kyung Hee University\\
\email{\{2007kkl, kdh2769, everyman123, hyoseok\}@khu.ac.kr}
}

\maketitle

\begin{abstract}

Accurate LiDAR-camera calibration is essential for robust multi-modal perception.
Targetless approaches avoid manual setup but remain limited by the scarcity of discriminative cross-modal features.
Recent methods address this by reconstructing the scene within a differentiable model, enabling extrinsic optimization through dense photometric supervision.
Among these, 3D Gaussian Splatting (3DGS) has been widely adopted as a geometric proxy that bridges LiDAR and camera within a single differentiable framework.
However, since 3DGS was originally designed for novel view synthesis, existing methods tend to prioritize rendering quality, causing the proxy geometry to drift from the true LiDAR structure.
We propose a framework that preserves the metric geometry of the Gaussian proxy by aggregating multi-view LiDAR observations for dense depth supervision and blocking photometric gradients from updating the Gaussian spatial parameters.
We validate our method on public driving datasets, where it consistently outperforms existing targetless methods in calibration accuracy.
\end{abstract}

\section{Introduction}
\label{sec:intro}

Autonomous systems, ranging from self-driving vehicles to mobile robots, rely on multi-modal sensor fusion for robust environmental perception~\cite{fan2025lidar, thrun2006stanley, caesar2020nuscenes, wang2023multi}. 
Among various sensor combinations, the integration of cameras and LiDAR is widely adopted, as cameras provide dense semantic information while LiDAR captures precise 3D geometric structures~\cite{vora2020pointpainting, guan2025lidar}. 
By combining these two modalities, perception systems can compensate for the limitations of each sensor, such as the scale ambiguity in monocular vision or the spatial sparsity of LiDAR data~\cite{ qiu2019deeplidar, song2024graphbev}. 
However, the effectiveness of this sensor fusion depends on accurate spatial alignment between the sensors, which is known as LiDAR-Camera extrinsic calibration.

For such spatial alignment, target-based methods~\cite{zhang2004extrinsic, zhou2018automatic, kim2024camera} have served as the standard approach. 
These methods rely on physical targets with known geometry, such as checkerboards, to establish precise correspondences between the two sensors. 
While target-based methods achieve high accuracy under controlled conditions, they require careful manual setup and a prepared environment, which limits their use in practice~\cite{yan2022opencalib}. 
In particular, recalibration during system operation is infeasible with this approach, making it difficult to correct sensor misalignment caused by mechanical vibrations or gradual drift in real-world deployments~\cite{lv2021lccnet, zhu2020online}.

Motivated by this overhead, targetless calibration methods have been actively explored~\cite{levinson2013automatic, zhang2021line, yuan2021pixel}. 
Rather than relying on artificial markers, these approaches leverage shared geometric primitives such as edges and structural lines that can be extracted from both LiDAR and camera data. 
However, the accuracy of conventional targetless methods is constrained by the scarcity of such discriminative features in typical driving scenes, especially in texture-poor or structurally repetitive environments~\cite{zhang2025higs}. 
To overcome this limitation, recent approaches have shifted from extracting individual features to representing the entire scene with a differentiable model, enabling extrinsic optimization through dense pixel-level supervision rather than sparse feature correspondences.

Neural Radiance Fields for View Synthesis (NeRF)~\cite{mildenhall2021nerf} exemplifies this direction, but the computational burden of volumetric rendering often makes NeRF-based calibration~\cite{zhou2023inf, yang2024unical} impractical for real-time or large-scale use.
Recently, 3D Gaussian Splatting (3DGS)~\cite{kerbl20233d} has emerged as an efficient scene representation that enables fast, differentiable rendering through explicit point-based primitives. Since each Gaussian can be initialized from a LiDAR point, inheriting its precise metric position, and then rendered onto the camera image plane through differentiable rasterization, 3DGS can effectively function as a geometric proxy that bridges the LiDAR coordinate system and the camera image domain.

This proxy property provides a clear advantage for LiDAR-Camera extrinsic calibration, and several recent methods have adopted 3DGS for this task~\cite{herau20243dgs,zhou2025robust,zhang2025higs, jung2026targetlesslidarcameracalibrationneural}. 
These methods share a similar pipeline in which Gaussians are initialized from LiDAR points, rendered into images through the Gaussian proxy, and jointly optimized with the extrinsic parameters by minimizing image-domain losses.
Since 3DGS was originally designed for novel view synthesis, its optimization is inherently driven toward improving rendering quality, and existing calibration methods naturally inherit this tendency.
However, this rendering-oriented optimization freely adjusts Gaussian positions and shapes to reduce photometric residuals, even at the cost of distorting the underlying geometric structure.
Although existing methods constrain the proxy geometry using sparse depth supervision from a single LiDAR scan at each timestep, this coverage is insufficient to prevent the proxy's geometric parameters from drifting away from the true LiDAR structure during joint optimization.
As a result, the calibration accuracy that depends on metric faithfulness deteriorates.
We refer to this phenomenon as \textit{Geometric Decay}, a gradual loss of metric faithfulness in the proxy geometry that degrades calibration accuracy and stability.

To address this problem, we propose \textbf{Geo}metry-\textbf{P}reserving \textbf{Calib}ration (GeoP-Calib), a framework that fully exploits the geometric proxy property of 3DGS for extrinsic calibration. 
First, we introduce Dense Depth Anchoring (DDA), which aggregates multi-view LiDAR observations to construct a dense depth prior. DDA increases the spatial coverage of depth supervision, providing a rigid metric foundation that penalizes geometric drift in regions where a single scan offers no constraint. 
To handle occlusion artifacts that arise from multi-view accumulation, DDA further introduces a Volumetric Soft Mask (VSM) that uses the rendered depth of the Gaussian proxy as a continuous visibility prior, down-weighting occluded LiDAR projections and resolving spatial ambiguities.
Second, we design Gradient Decoupling (GD), which blocks the backpropagation of photometric gradients to the Gaussian spatial parameters, specifically position and covariance. 
GD prevents texture-driven residuals from deforming the proxy geometry, effectively mitigating \textit{Geometric Decay} and preserving the bridge role of the proxy between LiDAR and camera. 

In summary, our key contributions are:
(i) We identify and analyze a photometric-geometric conflict in 3DGS-based calibration, and empirically demonstrate that photometric supervision can induce \textit{Geometric Decay} of the Gaussian proxy.
(ii) We propose GeoP-Calib, a novel framework that integrates Dense Depth Anchoring and Gradient Decoupling to mitigate \textit{Geometric Decay} of the Gaussian proxy.
(iii) Extensive evaluations on KITTI odometry~\cite{Geiger2012CVPR} and KITTI-360~\cite{Liao2021ARXIV} demonstrate consistent improvements in calibration accuracy, with particularly notable gains in translation estimation.

\section{Related Works}
\label{sec:related}

\subsection{Target-Based LiDAR-Camera Calibration}
Target-based methods estimate extrinsics by observing artificial objects with known geometry, such as checkerboards or fiducial markers, and solving for the transformation via explicit 2D-3D correspondences.
Zhang and Pless~\cite{zhang2004extrinsic} established the foundational point-to-plane formulation for LiDAR-Camera pairs, and Zhou~\etal~\cite{zhou2018automatic} relaxed the target constraints by exploiting multimodal edge line and plane correspondences. 
Another line of research, Beltr\'an~\etal~\cite{beltran2022automatic} diversified target-based calibration with circular targets, and Kim~\etal~\cite{kim2024camera} further improved it by more accurate circle-center localization.
Despite their precision, target-based methods require meticulous manual setup, limiting their scalability for deployments where sensor extrinsics drift over time.


\subsection{Targetless LiDAR-Camera Calibration}
Targetless methods have exploited naturally occurring scene structures, avoiding dedicated calibration objects.
One line of work aligned geometric features across modalities, such as edges~\cite{kang2020automatic, yuan2021pixel}, 3D-2D lines~\cite{zhang2021line}, and cross-modal structural correspondences~\cite{ou2023targetless}.
Another direction optimized photometric consistency directly via mutual information or intensity-based objectives~\cite{irie2016target, pandey2012automatic, koide2023general}.
On the learning side, Iyer~\etal~\cite{iyer2018calibnet} and Lv~\etal~\cite{lv2021lccnet} regressed extrinsics from raw LiDAR-image pairs using CNNs and cost volumes, while Zhang~\etal~\cite{zhang2025claim} more recently combined foundation-model-based monocular depth with LiDAR intensity alignment.
Despite their flexibility, these approaches remain limited by the inherent sparsity of LiDAR point clouds, which restricts both the availability of discriminative features and the reliability of cross-modal comparison~\cite{zhang2025higs}.

\subsection{Scene Reconstruction-Based LiDAR-Camera Calibration}
Scene reconstruction-based methods formulate calibration as a joint optimization of a differentiable scene representation and extrinsic parameters, where the photometric loss between rendered and observed images drives pose refinement.
NeRF-based approaches~\cite{zhou2023inf, herau2024soac} train neural radiance fields while jointly optimizing extrinsics, but the volumetric rendering pipeline incurs substantial computational cost, often requiring several hours per calibration.
To overcome this efficiency bottleneck, recent methods adopt 3DGS~\cite{kerbl20233d} as the underlying representation.
Herau~\etal~\cite{herau20243dgs} first applied 3DGS to LiDAR-Camera calibration by fixing Gaussians on LiDAR points and jointly optimizing the scene model and extrinsics.
Zhou~\etal~\cite{zhou2025robust} adopted 2D Gaussian Splatting~\cite{huang20242d} and refined extrinsics through reprojection and triangulation losses.
Zhang~\etal~\cite{zhang2025higs} decoupled scene modeling from extrinsic refinement via a hierarchical coarse-to-fine architecture.
Jung~\etal~\cite{jung2026targetlesslidarcameracalibrationneural} introduced fixed anchor Gaussians and learnable auxiliary Gaussians with a camera rig optimization strategy for consistent multi-view calibration.

\section{Problem Definition}
\label{sec:problem_definition}

\subsection{3DGS Preliminaries}
\label{sec:3dgs_prelim}

3DGS~\cite{kerbl20233d} represents a 3D scene using a set of anisotropic 3D Gaussians, $\mathcal{G}=\{G_i\}_{i=1}^{N}$. Each Gaussian $G_i$ is parameterized by its mean $\mu_i \in \mathbb{R}^3$, covariance $\Sigma_i \in \mathbb{R}^{3 \times 3}$, opacity $\alpha_i \in (0,1)$, and color coefficients $c_i$. The covariance matrix $\Sigma_i$ is typically decomposed into a rotation matrix $R_i$ and a scaling matrix $S_i$ as $\Sigma_i = R_i S_i S_i^\top R_i^\top$ to maintain its physical validity during optimization.

To render the scene from a specific viewpoint, the 3D Gaussians are projected onto the 2D image plane. Given World to Camera $\mathbf{T}_{cw} \in SE(3)$ and intrinsics $\mathbf{K}$, each 3D Gaussian is projected to form a 2D Gaussian density $G_i^{2D}(\mathbf{p})$ evaluated at pixel $\mathbf{p} \in \mathbb{R}^2$. 

3DGS utilizes tile-based rasterization with alpha-compositing to render images in a differentiable manner. For each pixel $\mathbf{p}$, the rendered color $\Phi_{color}$ and depth $\Phi_{depth}$ are computed by blending the $N$ Gaussians overlapping that pixel:
\begin{equation}
\Phi_{color}(\mathcal{G}, \mathbf{T}_{cw})(\mathbf{p}) = \sum_{i=1}^{N} T_i(\mathbf{p}) \alpha_i G_i^{2D}(\mathbf{p}) c_i,
\end{equation}
\begin{equation}
\Phi_{depth}(\mathcal{G}, \mathbf{T}_{cw})(\mathbf{p}) = \sum_{i=1}^{N} T_i(\mathbf{p}) \alpha_i G_i^{2D}(\mathbf{p}) d_i,
\end{equation}
where $T_i(\mathbf{p}) = \prod_{j=1}^{i-1} (1 - \alpha_j G_j^{2D}(\mathbf{p}))$ represents the accumulated transmittance, and $d_i$ denotes the depth of the $i$-th Gaussian in the camera coordinate system. The entire rendering pipeline is fully differentiable with respect to both the scene parameters $\mathcal{G}$ and the camera pose $\mathbf{T}_{cw}$, enabling gradient-based optimization for calibration tasks.

\subsection{3DGS-based Calibration Preliminaries}
The objective of 3DGS-based LiDAR-Camera calibration is to estimate the optimal scene representation and the extrinsic transformation by maximizing the joint posterior probability~\cite{zhang2025higs}. Given a set of camera images $\mathcal{I} = \{I^t\}$, a set of LiDAR point clouds $\mathcal{P} = \{\mathcal{P}^t\}$, and the world to LiDAR poses $\mathbf{T}_{lw} = \{\mathbf{T}_{lw}^t\}$, the optimization problem is formulated as:
\begin{equation}
    \mathcal{\hat{G}}, \mathbf{\hat{T}}_{cl} = \arg\max_{\mathcal{G}, \mathbf{T}_{cl}} P(\mathcal{G}, \mathbf{T}_{cl} \mid \mathcal{I}, \mathcal{P}, \mathbf{T}_{lw}),
\end{equation}
where $\mathcal{G}$ denotes the set of 3D Gaussian parameters, and $\mathbf{T}_{cl} \in SE(3)$ represents the extrinsic transformation from the LiDAR coordinate system ($l$) to the camera coordinate system ($c$), parameterized by a rotation matrix $\mathbf{R} \in SO(3)$ and a translation vector $\mathbf{t} \in \mathbb{R}^3$.

To solve this MAP estimation, existing frameworks typically minimize a multi-modal objective function. The extrinsic parameters are optimized through a differentiable rendering process $\Phi(\mathcal{G}, \mathbf{T}_{cw})$, where the camera pose at time $t$ is defined as a function of the calibration target: $\mathbf{T}_{cw}^t = \mathbf{T}_{cl} \cdot \mathbf{T}_{lw}^t$.

\noindent\textbf{Photometric Alignment.}
The first objective ensures that the rendered appearance matches the observed images. The photometric alignment loss $\mathcal{L}_{pho}$ is generally defined using a distance metric $\mathcal{D}_{photo}$:
\begin{equation}
\label{eq:photo_loss}
    \mathcal{L}_{pho} = \mathcal{D}_{photo} \Big( I^t, \, \Phi_{color}(\mathcal{G}, \mathbf{T}_{cl} \mathbf{T}_{lw}^t) \Big),
\end{equation}
where $\mathcal{D}_{photo}$ measures the visual discrepancy. In practice, this metric typically employs the pixel-wise $L_1$ distance to ensure absolute color alignment and minimize rendering residuals across the observed views~\cite{zhang2025higs,zhou2025robust,herau20243dgs,jung2026targetlesslidarcameracalibrationneural}.

\noindent\textbf{Cross-view Geometric Consistency.}
Recent works further regularize the extrinsics using reprojection consistency~\cite{zhou2025robust,zhang2025higs}. Utilizing the rendered depth map $D^t = \Phi_{depth}(\mathcal{G}, \mathbf{T}_{cl} \mathbf{T}_{lw}^t)$, a pixel $\mathbf{p}^t$ in the source frame $t$ can be warped to the target frame $t+s$ as follows:
\begin{equation}
    \tilde{\mathbf{p}}^{t+s} = \pi \left( \mathbf{K} (\mathbf{T}_{cl} \mathbf{T}_{lw}^{t+s}) (\mathbf{T}_{cl} \mathbf{T}_{lw}^{t})^{-1} \big( D^t(\mathbf{p}^t) \mathbf{K}^{-1} \bar{\mathbf{p}}^t \big) \right),
\end{equation}
where $\mathbf{K}$ is the camera intrinsic matrix, $\bar{\mathbf{p}}^t$ is the homogeneous coordinate of $\mathbf{p}^t$, and $\pi(\cdot)$ denotes the perspective projection function that divides by the depth component. The reprojection loss $\mathcal{L}_{rep}$ minimizes the intensity discrepancy between the warped and target pixels:
\begin{equation}
\label{eq:rep_loss}
    \mathcal{L}_{rep} = \mathcal{D}_{rep} \Big( I^{t+s}(\tilde{\mathbf{p}}^{t+s}), \, I^t(\mathbf{p}^t) \Big),
\end{equation}
where $\mathcal{D}_{rep}$ represents a photometric distance function.


\noindent\textbf{Optimization through Gaussian Proxy.}
In 3DGS-based calibration, the extrinsic $\mathbf{T}_{cl}$ is refined by minimizing image-domain objectives such as photometric alignment and cross-view reprojection consistency. We define the image-domain objective as
\begin{equation}
\mathcal{L}_{img} \;=\; \mathcal{L}_{pho} \;+\; \lambda_{rep}\mathcal{L}_{rep}.
\end{equation}
Since both $\mathcal{L}_{pho}$ and $\mathcal{L}_{rep}$ are computed from rendered outputs, their gradients with respect to $\mathbf{T}_{cl}$ flow through the differentiable renderer:
\begin{equation}
\frac{\partial \mathcal{L}_{img}}{\partial \mathbf{T}_{cl}}
=
\frac{\partial \mathcal{L}_{img}}{\partial \Phi(\mathcal{G}, \mathbf{T}_{cw})}
\cdot
\frac{\partial \Phi(\mathcal{G}, \mathbf{T}_{cw})}{\partial \mathbf{T}_{cl}},
\end{equation}
where $\mathbf{T}_{cw}=\mathbf{T}_{cl}\mathbf{T}_{lw}$.
Importantly, the renderer Jacobian $\partial \Phi(\mathcal{G}, \mathbf{T}_{cw})/\partial \mathbf{T}_{cl}$ depends on the current proxy state $\mathcal{G}$, implying that the reliability of the extrinsic update is tightly coupled to the geometric faithfulness of $\mathcal{G}$ with respect to the underlying LiDAR structure $\mathcal{P}$.

\noindent\textbf{LiDAR Depth Anchoring.}
To explicitly constrain the proxy geometry in metric space, prior 3DGS-based calibration pipelines commonly include a LiDAR-derived depth anchoring term~\cite{zhang2025higs, zhou2025robust}.
In the standard setting, this term is instantiated from a single LiDAR scan at time $t$ by projecting LiDAR points onto the image plane and comparing the rendered depth with the LiDAR depth at valid projection pixels.
Since the extrinsic translation is still uncertain during optimization, the depth rendering adopts a rotation-only extrinsic $\mathbf{\bar{T}}_{cl}=[\mathbf{R}_{cl}\mid\mathbf{0}]$, isolating the depth supervision from the translation uncertainty by keeping both measurements at the LiDAR origin.
The depth anchoring loss is then formulated as:
\begin{equation}
\label{eq:sparse_depth_loss}
\mathcal{L}^{sparse}_{depth}
=
\frac{1}{|\Omega^t|}
\sum_{\mathbf{p}\in\Omega^t}
\mathcal{D}_{depth}\!\left( D_{rend}(\mathbf{p}),\, D_{l}(\mathbf{p}) \right),
\end{equation}
where $\Omega^t$ denotes the set of valid projected pixels, $D_{rend}=\Phi_{depth}(\mathcal{G},\mathbf{\bar{T}}_{cl}\mathbf{T}_{lw}^{t})$ is the depth rendered from the virtual viewpoint at the LiDAR origin with the camera orientation, and $D_{l}$ is the depth obtained from the projected LiDAR points.
We denote $\mathcal{L}_{depth}=\mathcal{L}^{sparse}_{depth}$ to represent the depth objective in existing baselines.

\noindent\textbf{Overall Objective.}
The full objective minimized in typical 3DGS-based calibration pipelines is a weighted sum of the photometric alignment, cross-view consistency, and geometric constraints:
\begin{equation}
\label{eq:total_loss}
    \mathcal{L}_{total} = \mathcal{L}_{pho} + \lambda_{rep}\mathcal{L}_{rep} + \lambda_{depth}\mathcal{L}_{depth},
\end{equation}
where $\lambda_{rep}$ and $\lambda_{depth}$ are balancing weights.


\subsection{Limitations of Sparse LiDAR Supervision}
\label{sec:limitation_sparse}

As Eq.~\ref{eq:sparse_depth_loss} supervises depth only at pixels where LiDAR points project, a single scan provides supervision over only a small fraction of the image, leaving most regions unconstrained in metric space.
Notably, rendered depth is a view-dependent output produced at rendering time, $D_{rend}=\Phi_{depth}(\mathcal{G}, \bar{\mathbf{T}}_{cl}\mathbf{T}_{lw}^t)$.
Thus, outside the supervised set $\Omega^t$, the proxy geometry can change without being directly penalized by the depth objective.
As a result, many Gaussians can drift in underconstrained areas while still producing plausible renderings, which undermines the geometric faithfulness of the proxy and can mislead extrinsic updates.

\subsection{Analysis of Photometric-Geometric Conflict}
\label{sec:geo_conflict}

\begin{figure*}[!t]
    \centering
    
    \begin{minipage}[c]{0.43\textwidth}
        \begin{subfigure}{\linewidth}
            \centering
            \includegraphics[width=\linewidth]{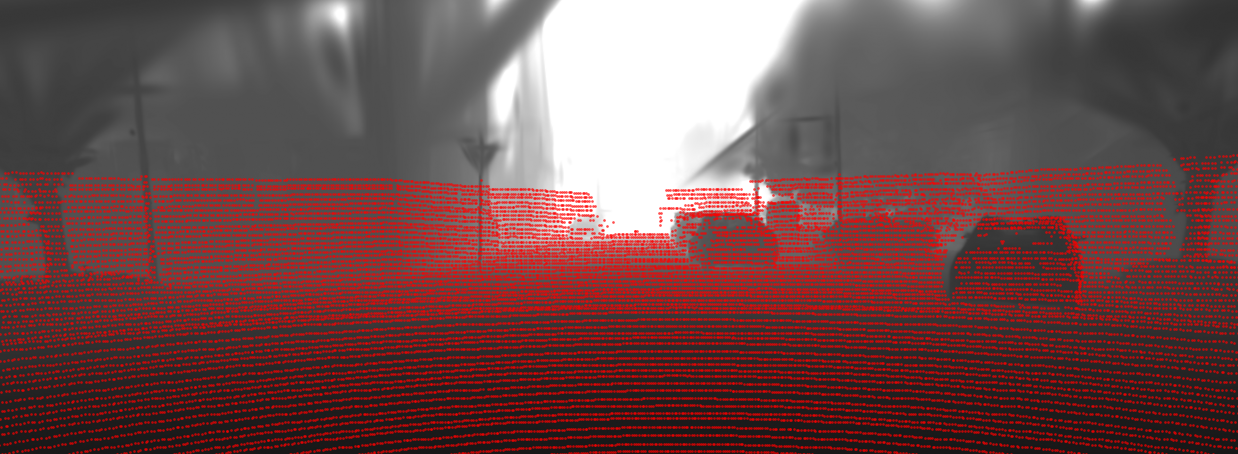}
            \caption{}
            \label{fig:geo_decay:a}
        \end{subfigure}
        
        \begin{subfigure}{\linewidth}
            \centering
            \includegraphics[width=\linewidth]{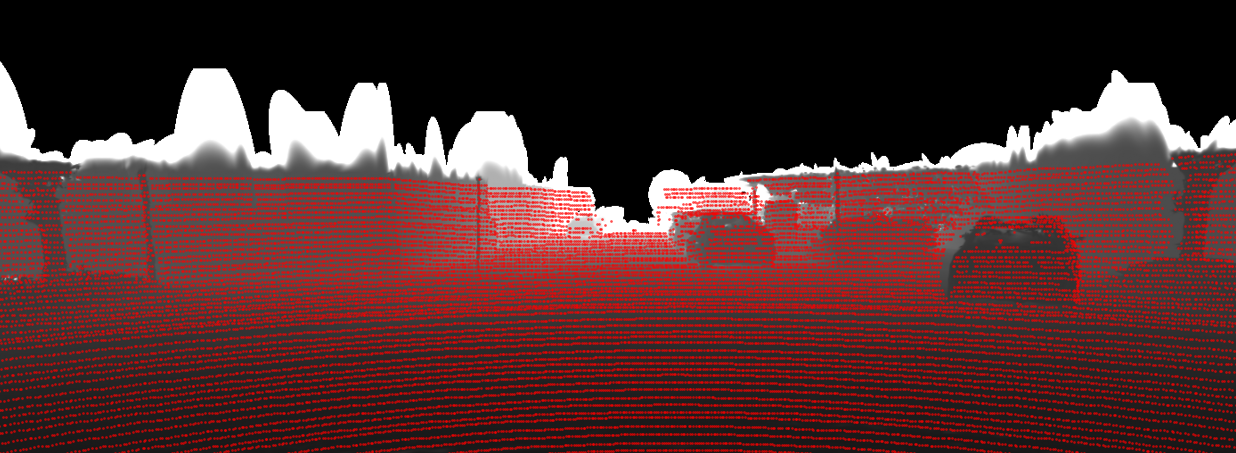}
            \caption{}
            \label{fig:geo_decay:b}
        \end{subfigure}
    \end{minipage}
    \hfill 
    \begin{subfigure}[c]{0.55\textwidth}
        \centering
        \includegraphics[width=1.0\linewidth]{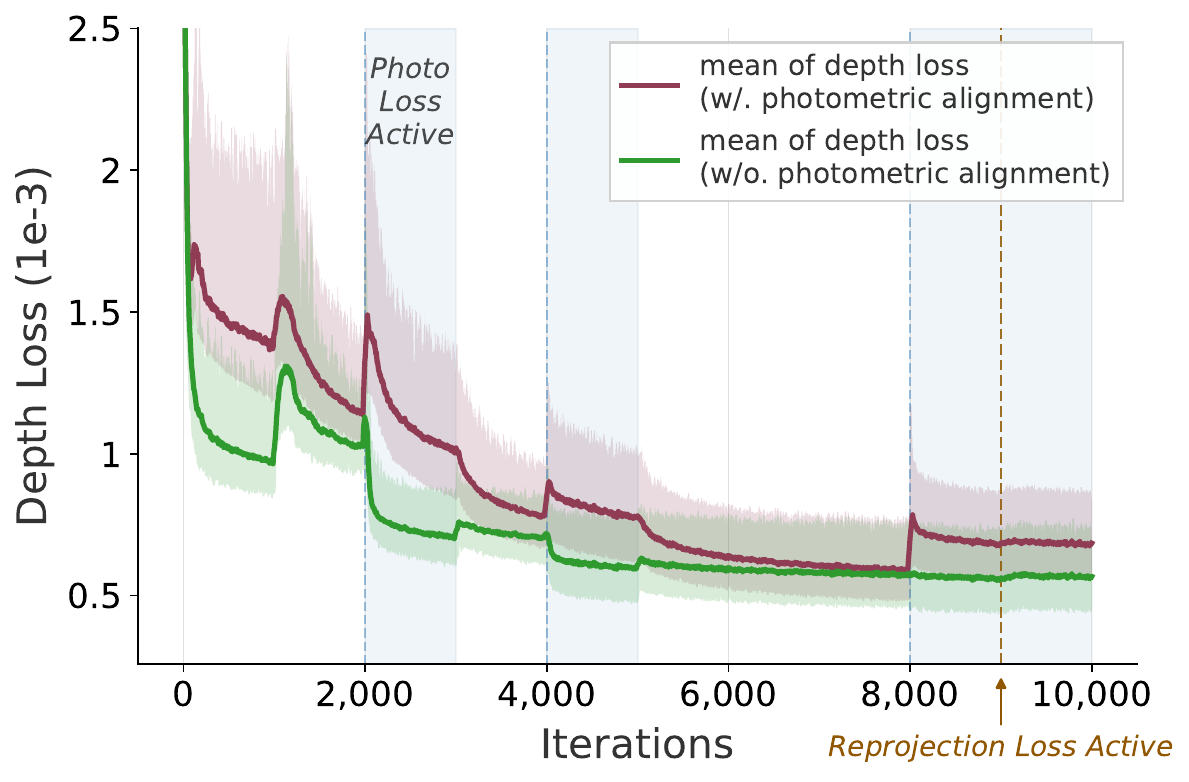}
        \caption{}
        \label{fig:geo_decay:c}
    \end{subfigure}

    \caption{
        Illustration of Geometric Decay:
        (a) Rendered depth map of 3DGS optimized using Base pipeline with photometric alignment loss.
        (b) Rendered depth map of 3DGS optimized with only depth loss.
        (c) Evolution of the depth loss on KITTI-360 Seq.~1. Shaded regions indicate min–max. Reprojection is used when photometric alignment is disabled, and both are enabled after 9000 iterations.
    }
    \label{fig:photometric_geometric_conflict}
\end{figure*}

In 3DGS-based LiDAR-Camera calibration, the extrinsic parameters are optimized through image-domain objectives in Eq.~\ref{eq:photo_loss} and Eq.~\ref{eq:rep_loss}, and the resulting gradients backpropagate through the Gaussian proxy.
Many pipelines intentionally limit Gaussian density, for instance, by tying the primitive budget to LiDAR voxelization or point sampling. \cite{zhang2025higs, zhou2025robust, herau20243dgs,jung2026targetlesslidarcameracalibrationneural}
This constraint is designed to preserve metric correspondence with the LiDAR initialization and prevent the model from introducing excessive geometric degrees of freedom to compensate for photometric errors. \cite{herau20243dgs}
Under this constrained proxy resolution, photometric observations containing fine textures cannot always be fully explained by appearance parameters alone.
Consequently, the optimizer may reduce photometric error by modifying geometric parameters such as Gaussian means and covariances, trading metric geometric fidelity for visually plausible renderings.
We refer to this texture-driven distortion of the proxy geometry as \textit{Geometric Decay}, where the Gaussian metric structure loses faithfulness to the LiDAR points.

Fig.~\ref{fig:photometric_geometric_conflict} illustrates this phenomenon.
Fig.~\ref{fig:geo_decay:a} shows rendered depth appearing outside LiDAR-supported regions, whereas Fig.~\ref{fig:geo_decay:b} keeps rendered depth more localized to regions where LiDAR points exist.
Fig.~\ref{fig:geo_decay:c} plots the depth loss, defined as an $L_1$ discrepancy between rendered depth and LiDAR depth in the inverse-depth domain. 
The red curve (with photometric alignment) shows a higher mean loss and larger variability, particularly during the shaded interval when photometric alignment is enabled, compared to the run without photometric alignment.
These results indicate that optimizing photometric consistency can degrade the geometric accuracy required for LiDAR-Camera calibration.

\begin{figure}[t]
\centering
\includegraphics[width=\linewidth]{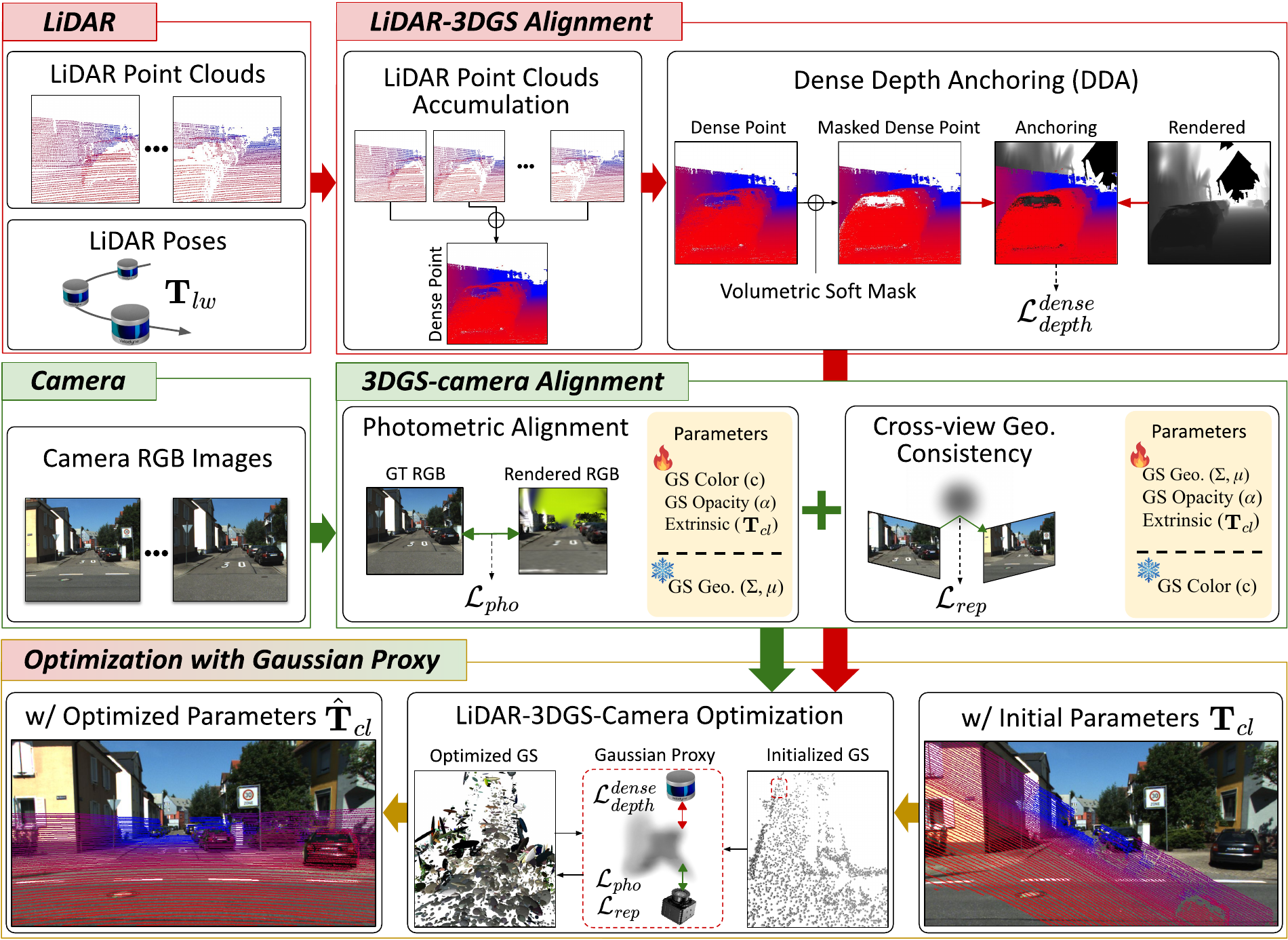}
\caption{Overview of the proposed geometry-preserving calibration integration.}
\label{fig:pipeline}
\end{figure}

\section{Geometry-Preserving LiDAR-Camera Calibration}


We propose Geometry-Preserving LiDAR-Camera Calibration (GeoP-Calib), a framework that preserves LiDAR-consistent proxy geometry throughout joint optimization. 
GeoP-Calib addresses the conflict identified in Sec.~\ref{sec:problem_definition} through two complementary components: (1) Dense Depth Anchoring (DDA), which establishes a rigid metric foundation by aggregating multi-view LiDAR observations, and (2) Gradient Decoupling (GD), which blocks the gradients of the photometric alignment loss $\mathcal{L}_{pho}$ from updating the Gaussian spatial parameters, thereby mitigating \textit{Geometric Decay}.
These two components operate cooperatively while keeping their gradient effects decoupled.
DDA governs the spatial parameters of the proxy, while GD allows photometric alignment to refine only appearance and sensor extrinsics without interfering with the established geometry.

\subsection{Dense Depth Anchoring with Volumetric Masking}

To resolve the structural instability caused by the sparsity of single-scan LiDAR data, we propose \textbf{Dense Depth Anchoring} (DDA). This module provides a continuous geometric reference by accumulating multi-view LiDAR observations, ensuring that the Gaussian proxy remains rigidly aligned with the global scene structure.

\noindent\textbf{Geometric Foundation via Point Accumulation.}
As identified in Sec.~\ref{sec:limitation_sparse}, the lack of spatial constraints in unsupervised void regions leads to Gaussian drift. We mitigate this by constructing a dense global point cloud $\mathcal{P}_{global} = \bigcup_{t} (\mathbf{T}_{lw}^t)^{-1} \mathcal{P}^t$. By projecting $\mathcal{P}_{global}$ from the LiDAR viewpoint onto the image plane, we generate a dense depth prior $D^{dense}_{l}$ that covers a significantly larger portion of the image space than a single scan. This dense anchor effectively locks the spatial parameters ($\mu, \Sigma$) of the Gaussians, preventing them from drifting even in regions where the current frame's LiDAR data is absent.



\noindent\textbf{Volumetric Soft Mask for Occlusion Handling.}
A fundamental challenge in dense point accumulation is the \textit{occlusion ambiguity}, where points from distant viewpoints are incorrectly projected onto foreground objects. To effectively resolve this, we leverage the volumetric nature of 3DGS. Since multi-view reprojection evaluates structural consistency across different camera poses, it naturally encourages Gaussians to accurately fill valid spatial regions. This inherent spatial awareness makes the splatted Gaussians a useful geometric cue for reasoning about occlusions.


Building upon this insight, we propose utilizing the 3DGS representation itself as a continuous occlusion filter, introducing a \textbf{Volumetric Soft Mask} (VSM). Because the rendered depth $D_{rend}$ represents the expected ray termination, it implicitly captures the visibility state of the scene. We formulate the visibility weight $W_{vis}$ to adaptively filter the accumulated points as follows:
\begin{equation}
\label{eq:vsm}
    W_{vis}(\mathbf{p}) = \sigma \left( \beta \cdot \big( D_{rend}(\mathbf{p})\,(1+\tau) - D_{l}^{dense}(\mathbf{p}) \big) \right),
\end{equation}
where $\sigma(\cdot)$ is the sigmoid function, $\tau$ is a depth-proportional tolerance margin, and $\beta$ controls the sharpness of the transition. By smoothly penalizing points located significantly behind the rendered Gaussian surface ($D_{l}^{dense} \gg D_{rend}$), this soft masking strategy effectively neutralizes invalid projections while maintaining a stable, differentiable loss landscape.

\noindent\textbf{Dense Depth Consistency Loss.}
Using the accumulated points projected onto the current view, we explicitly enforce consistency between the rendered depth and the LiDAR-implied depth at the corresponding pixel locations. Let $\Omega^{dense}$ denote the set of valid projected LiDAR points, and let $M(\mathbf{p})\in\{0,1\}$ be a binary validity mask indicating whether the rendered depth is defined at pixel $\mathbf{p}$. We compute the depth residual in the inverse-depth domain to provide a more stable parameterization across a wide depth range~\cite{civera2008inverse}:
\begin{equation}
    \Delta d^{inv}(\mathbf{p}) = \left| d^{inv}_{rend}(\mathbf{p}) - d^{inv}_{l}(\mathbf{p}) \right|,
    \quad
    d^{inv}_{l}(\mathbf{p}) = \frac{1}{D_{l}^{dense}(\mathbf{p})+\epsilon},
\end{equation}
where $d^{inv}_{rend}(\mathbf{p})$ denotes the rendered inverse depth at $\mathbf{p}$, obtained using the same rotation-only extrinsic $\bar{\mathbf{T}}_{cl}$ as in Eq.~\ref{eq:sparse_depth_loss}.
The final dense depth anchoring loss combines the validity mask and the soft visibility weight:
\begin{equation}
    \mathcal{L}^{dense}_{depth}
    =
    \frac{\sum_{\mathbf{p}\in \Omega^{dense}} M(\mathbf{p})\,W_{vis}(\mathbf{p})\,\Delta d^{inv}(\mathbf{p})}
    {\sum_{\mathbf{p}\in \Omega^{dense}} M(\mathbf{p})\,W_{vis}(\mathbf{p})+\epsilon}.
\end{equation}

\subsection{Gradient Decoupling for Structural Protection}

To mitigate the \textit{Geometric Decay} identified in Sec.~\ref{sec:geo_conflict}, we introduce \textbf{Gradient Decoupling} (GD). 
GD blocks the backpropagation of the photometric alignment loss to the Gaussian spatial parameters, preventing texture-driven photometric residuals from directly deforming the established metric structure.

\noindent\textbf{Implementation via Gradient Blocking.}
In a standard joint optimization, the photometric alignment loss $\mathcal{L}_{pho}$ backpropagates through the entire set of Gaussian parameters $\mathcal{G}$, comprising $\{\mu, \Sigma, \alpha, c\}$.
To protect the structural integrity, we apply a stop-gradient operator $\text{sg}[\cdot]$ to the spatial parameters $(\mu, \Sigma)$ specifically within the color rendering process. The modified photometric objective is defined as:
\begin{equation}
    \mathcal{L}_{pho}^{decoupled} = \mathcal{D}_{photo} \Big( I^t,\, \Phi_{color}(\text{sg}[\mu, \Sigma], \alpha, c, \mathbf{T}_{cl} \cdot \mathbf{T}_{lw}^t) \Big).
\end{equation}
By explicitly decoupling these gradients, we ensure that the refinement of scene appearance and sensor extrinsics proceeds without inducing non-physical deformations in the underlying geometry.

\noindent\textbf{Selective Gradient Blocking.}
As summarized in Fig.~\ref{fig:pipeline}, we apply GD only to $\mathcal{L}_{pho}$, while allowing the reprojection loss $\mathcal{L}_{rep}$ to update $(\mu,\Sigma)$.
Although $\mathcal{L}_{rep}$ also uses a photometric distance, it compares ground-truth image intensities through depth-induced reprojection correspondences without involving the learned Gaussian color attributes. 
As a result, its gradients to $(\mu,\Sigma)$ are mediated by the reprojection geometry and primarily encourage view-consistent structure rather than per-view texture fitting.
This is empirically supported by Fig.~\ref{fig:geo_decay:c}. 
The depth loss sharply increases once the photometric alignment loss is activated, yet the subsequent activation of the reprojection loss does not induce a comparable degradation.
This design leverages image-domain supervision while mitigating \textit{Geometric Decay} and preserving LiDAR-anchored structure during extrinsic refinement.

Combining both components, GeoP-Calib modifies the baseline objective (Eq.~\ref{eq:total_loss}) as:
\begin{equation}
    \mathcal{L}_{total}^{GP} = \mathcal{L}_{pho}^{decoupled} + \lambda_{rep}\mathcal{L}_{rep} + \lambda_{depth}^{sparse}\mathcal{L}_{depth}^{sparse} + \lambda_{depth}^{dense}\mathcal{L}_{depth}^{dense}.
\end{equation}

\section{Experiment}
\label{sec:experiment}

\subsection{Experiment Setup}
\label{sec:experiment_setup}

\noindent\textbf{Dataset.}
We evaluated GeoP-Calib on KITTI-360~\cite{Liao2021ARXIV} and KITTI odometry~\cite{Geiger2012CVPR} (denoted as KITTI hereafter). For KITTI-360, we used the same five sequences as in~\cite{herau20243dgs, zhang2025higs}. For KITTI, we used subsets of sequences 5, 6, 7, 9, and 10 following~\cite{zhou2025robust}. On KITTI-360, calibration was run on the two forward-facing cameras and averaged. On KITTI, we used only cam2.

\noindent\textbf{Baselines and Protocols.}
We compared against GST~\cite{koide2023general} as a representative traditional optimization-based method, CLAIM~\cite{zhang2025claim} as a foundation-model-based method, and RobustCalib~\cite{zhou2025robust} and HiGS-Calib~\cite{zhang2025higs} as recent 3DGS-based methods. 
All methods shared a unified initialization where the camera pose was aligned with the forward-facing LiDAR frame, except GST, whose SuperGlue~\cite{sarlin20superglue} based correspondence initialization could fail under sparse LiDAR projections and lead to invalid or unstable optimization. 
Therefore, we initialized GST by perturbing the ground-truth extrinsics with a $1^\circ$ rotation and $0.2$\,m translation to ensure that it entered a valid optimization regime.

\begin{table*}[t]
\caption{Quantitative comparison of extrinsic calibration accuracy on KITTI and KITTI-360 datasets. Best results are in bold, and second-best results are underlined. Values in parentheses denote the standard deviation.}
\label{tab:calibration_errors}
\centering
\resizebox{\textwidth}{!}{%
\setlength{\tabcolsep}{2pt} 
\begin{tabular}{@{} c c *{10}{w{c}{1.05cm}} @{}}
\toprule
\multirow{2}{*}{\rotatebox{90}{}} & \multirow{2}{*}{Seq.}
& \multicolumn{2}{c}{\makebox[0pt]{GST~\cite{koide2023general}}}
& \multicolumn{2}{c}{\makebox[0pt]{Claim~\cite{zhang2025claim}}}
& \multicolumn{2}{c}{\makebox[0pt]{RobustCalib~\cite{zhou2025robust}}}
& \multicolumn{2}{c}{\makebox[0pt]{HiGS-Calib~\cite{zhang2025higs}}}
& \multicolumn{2}{c}{\makebox[0pt]{GeoP-Calib}} \\
\cmidrule(lr){3-4}\cmidrule(lr){5-6}\cmidrule(lr){7-8}\cmidrule(lr){9-10}\cmidrule(lr){11-12}
& & $E_r (^\circ)$ & $E_t (\text{m})$
  & $E_r (^\circ)$ & $E_t (\text{m})$
  & $E_r (^\circ)$ & $E_t (\text{m})$
  & $E_r (^\circ)$ & $E_t (\text{m})$
  & $E_r (^\circ)$ & $E_t (\text{m})$ \\
\midrule
\midrule

\multirow{12}{*}{\rotatebox{90}{KITTI-360}}
& \multirow{2}{*}{seq1}
  & 0.351 & 0.326  & 0.663 & 0.142  & 0.202 & 0.059  & \best{0.111} & \second{0.036}  & \second{0.124} & \best{0.031} \\
& & {\scriptsize(0.080)} & {\scriptsize(0.018)}
    & {\scriptsize(0.409)} & {\scriptsize(0.148)}
    & {\scriptsize(0.021)} & {\scriptsize(0.006)}
    & {\scriptsize(0.016)} & {\scriptsize(0.006)}
    & {\scriptsize(0.046)} & {\scriptsize(0.006)} \\

& \multirow{2}{*}{seq2}
  & 4.631 & 0.394  & 0.614 & \best{0.056}  & 0.272 & 0.107  & \second{0.177} & 0.087  & \best{0.145} & \second{0.059} \\
& & {\scriptsize(2.543)} & {\scriptsize(0.254)}
    & {\scriptsize(0.219)} & {\scriptsize(0.030)}
    & {\scriptsize(0.046)} & {\scriptsize(0.019)}
    & {\scriptsize(0.031)} & {\scriptsize(0.004)}
    & {\scriptsize(0.020)} & {\scriptsize(0.007)} \\

& \multirow{2}{*}{seq3}
  & 0.991 & 0.335  & 0.232 & 0.073  & 0.307 & \second{0.076}  & \second{0.103} & 0.177  & \best{0.074} & \best{0.059} \\
& & {\scriptsize(0.199)} & {\scriptsize(0.212)}
    & {\scriptsize(0.073)} & {\scriptsize(0.041)}
    & {\scriptsize(0.047)} & {\scriptsize(0.002)}
    & {\scriptsize(0.096)} & {\scriptsize(0.106)}
    & {\scriptsize(0.030)} & {\scriptsize(0.002)} \\

& \multirow{2}{*}{seq4}
  & 1.011 & 0.316  & 0.809 & 0.296  & 0.246 & 0.133  & \second{0.167} & \second{0.118}  & \best{0.110} & \best{0.071} \\
& & {\scriptsize(0.176)} & {\scriptsize(0.210)}
    & {\scriptsize(0.641)} & {\scriptsize(0.253)}
    & {\scriptsize(0.068)} & {\scriptsize(0.005)}
    & {\scriptsize(0.161)} & {\scriptsize(0.111)}
    & {\scriptsize(0.042)} & {\scriptsize(0.001)} \\

& \multirow{2}{*}{seq5}
  & 0.751 & 0.206  & 0.284 & \best{0.079}  & 0.248 & 0.109  & \best{0.141} & 0.098  & \second{0.153} & \second{0.093} \\
& & {\scriptsize(0.423)} & {\scriptsize(0.071)}
    & {\scriptsize(0.110)} & {\scriptsize(0.039)}
    & {\scriptsize(0.003)} & {\scriptsize(0.012)}
    & {\scriptsize(0.010)} & {\scriptsize(0.001)}
    & {\scriptsize(0.030)} & {\scriptsize(0.003)} \\

\cmidrule{2-12}
& \multirow{2}{*}{Avg.}
  & 1.547 & 0.315  & 0.520 & 0.129  & 0.255 & \second{0.097}  & \second{0.140} & 0.103  & \best{0.121} & \best{0.063} \\
& & {\scriptsize(1.570)} & {\scriptsize(0.099)}
    & {\scriptsize(0.303)} & {\scriptsize(0.126)}
    & {\scriptsize(0.055)} & {\scriptsize(0.028)}
    & {\scriptsize(0.059)} & {\scriptsize(0.055)}
    & {\scriptsize(0.041)} & {\scriptsize(0.021)} \\

\midrule

\multirow{12}{*}{\rotatebox{90}{KITTI}}
& \multirow{2}{*}{seq1}
  & 2.626 & 0.251  & 0.247 & 0.044  & 0.305 & 0.050  & \second{0.112} & \second{0.042}  & \best{0.105} & \best{0.034} \\
& & {\scriptsize(1.151)} & {\scriptsize(0.082)}
    & {\scriptsize(0.125)} & {\scriptsize(0.016)}
    & {\scriptsize(0.051)} & {\scriptsize(0.006)}
    & {\scriptsize(0.008)} & {\scriptsize(0.002)}
    & {\scriptsize(0.009)} & {\scriptsize(0.001)} \\

& \multirow{2}{*}{seq2}
  & 1.310 & 0.232  & 0.371 & 0.092  & 0.482 & 0.074  & \second{0.201} & \second{0.056}  & \best{0.160} & \best{0.054} \\
& & {\scriptsize(0.294)} & {\scriptsize(0.058)}
    & {\scriptsize(0.107)} & {\scriptsize(0.030)}
    & {\scriptsize(0.417)} & {\scriptsize(0.014)}
    & {\scriptsize(0.003)} & {\scriptsize(0.001)}
    & {\scriptsize(0.010)} & {\scriptsize(0.001)} \\

& \multirow{2}{*}{seq3}
  & 1.106 & 0.167  & \best{0.219} & \second{0.052}  & 0.300 & 0.083  & 0.239 & \second{0.052}  & \second{0.224} & \best{0.023} \\
& & {\scriptsize(0.678)} & {\scriptsize(0.050)}
    & {\scriptsize(0.066)} & {\scriptsize(0.017)}
    & {\scriptsize(0.018)} & {\scriptsize(0.008)}
    & {\scriptsize(0.016)} & {\scriptsize(0.005)}
    & {\scriptsize(0.018)} & {\scriptsize(0.001)} \\

& \multirow{2}{*}{seq4}
  & 0.512 & 0.413  & 0.354 & 0.067  & 0.298 & 0.084  & \second{0.269} & \second{0.062}  & \best{0.250} & \best{0.052} \\
& & {\scriptsize(0.186)} & {\scriptsize(0.159)}
    & {\scriptsize(0.078)} & {\scriptsize(0.018)}
    & {\scriptsize(0.102)} & {\scriptsize(0.009)}
    & {\scriptsize(0.011)} & {\scriptsize(0.001)}
    & {\scriptsize(0.024)} & {\scriptsize(0.001)} \\

& \multirow{2}{*}{seq5}
  & 2.801 & 0.327  & 0.411 & \best{0.051}  & 0.375 & 0.119  & \second{0.209} & 0.065  & \best{0.201} & \second{0.056} \\
& & {\scriptsize(1.475)} & {\scriptsize(0.178)}
    & {\scriptsize(0.056)} & {\scriptsize(0.013)}
    & {\scriptsize(0.150)} & {\scriptsize(0.012)}
    & {\scriptsize(0.007)} & {\scriptsize(0.003)}
    & {\scriptsize(0.016)} & {\scriptsize(0.002)} \\

\cmidrule{2-12}
& \multirow{2}{*}{Avg.}
  & 1.671 & 0.278  & 0.352 & 0.082  & 0.352 & 0.082  & \second{0.206} & \second{0.055}  & \best{0.188} & \best{0.044} \\
& & {\scriptsize(0.892)} & {\scriptsize(0.085)}
    & {\scriptsize(0.074)} & {\scriptsize(0.022)}
    & {\scriptsize(0.071)} & {\scriptsize(0.022)}
    & {\scriptsize(0.053)} & {\scriptsize(0.008)}
    & {\scriptsize(0.051)} & {\scriptsize(0.013)} \\

\bottomrule
\end{tabular}
}
\end{table*}

\noindent\textbf{Evaluation Metrics.}
We quantitatively evaluated calibration accuracy using the translation error $E_t$ and the geodesic rotation error $E_r$:
\begin{equation}
\begin{aligned}
E_t &= \left\lVert \mathbf{\hat{t}} - \mathbf{t^*} \right\rVert_2,
\qquad
E_r = \arccos\!\left( \frac{\mathrm{Tr}\!\left( (\mathbf{R^*})^\top \mathbf{\hat{R}} \right) - 1}{2} \right),
\end{aligned}
\end{equation}
where $(\mathbf{R^*}, \mathbf{t^*})$ and $(\mathbf{\hat{R}}, \mathbf{\hat{t}})$ denote the ground-truth and estimated extrinsics, respectively. We report $E_r$ in degrees by converting radians to degrees.



\subsection{Implementation Details}
\label{sec:imp_detail}
We built our method on top of the HiGS code base~\cite{zhang2025higs}.
Notably, GD is applied throughout all stages to prevent photometric residuals from distorting the scene geometry. 
We adopt a two-stage optimization schedule: a geometric warm-up stage using sparse depth supervision to establish stable coarse geometry, followed by a refinement stage that activates the dense depth anchoring loss $\mathcal{L}^{dense}_{depth}$. 
For detailed hyperparameter settings, we refer the reader to the supplementary material.

\begin{table}[t]
    \centering
    \small
    \renewcommand{\arraystretch}{0.8}
    \setlength{\tabcolsep}{7pt}
    
    \caption{Component analysis for GeoP-Calib on KITTI-360. Note that VSM is a sub-component of DDA and is not defined without it.}
    \label{tab:ablation_component}
    
    \begin{tabular}{@{}ccc ccc@{}}
        \toprule
        GD & DDA & VSM & $E_r (^\circ)$ & $E_t (\text{m})$ & $T (\text{s})$ \\
        \midrule
        \midrule
        
        & & 
        & 0.122 {\scriptsize(0.031)} & 0.077 {\scriptsize(0.022)} 
        & 875 {\scriptsize(16)} \\
        
        & \checkmark & \checkmark
        & 0.117 {\scriptsize(0.044)} & 0.068 {\scriptsize(0.021)} 
        & 923 {\scriptsize(16)} \\
        
        \checkmark & & 
        & 0.127 {\scriptsize(0.042)} & 0.067 {\scriptsize(0.022)} 
        & 866 {\scriptsize(21)} \\
        
        \checkmark & \checkmark & 
        & 0.124 {\scriptsize(0.045)} & 0.066 {\scriptsize(0.024)} 
        & -- \\

        \checkmark & \checkmark & \checkmark 
        & 0.121 {\scriptsize(0.041)} & 0.063 {\scriptsize(0.021)} 
        & 922 {\scriptsize(21)} \\
        
        \bottomrule
    \end{tabular}
\end{table}

\begin{figure}[t]
  \centering
  \begin{subfigure}[b]{0.49\linewidth}
    \centering
    \includegraphics[width=\linewidth]{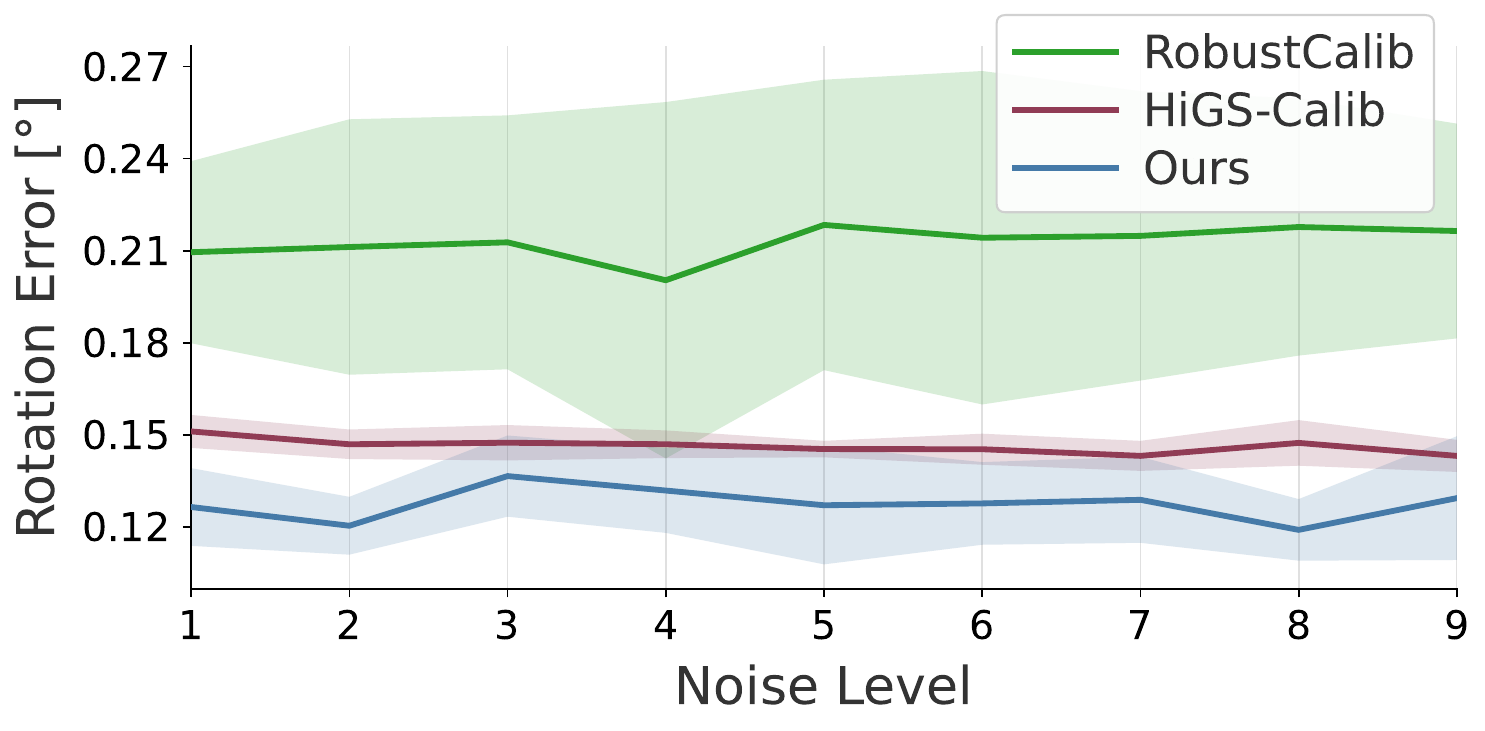}
    \caption{}
    \label{fig:noise-rotation}
  \end{subfigure}
  \hfill
  \begin{subfigure}[b]{0.48\linewidth}
    \centering
    \includegraphics[width=\linewidth]{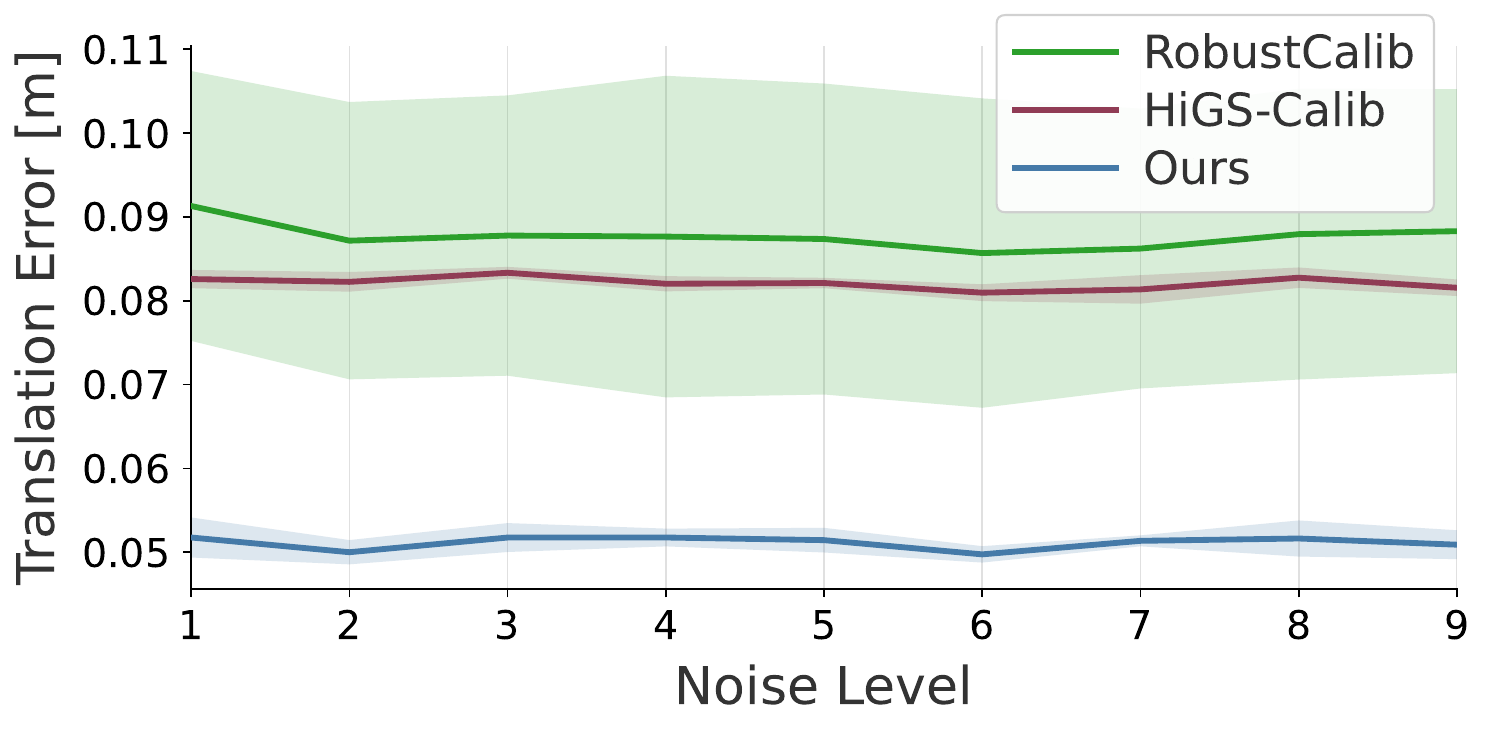}
    \caption{}
    \label{fig:noise-translation}
  \end{subfigure}
  \caption{Calibration error under varying noise levels: (\subref{fig:noise-rotation}) rotation and (\subref{fig:noise-translation}) translation.}
  \label{fig:noise-experiment}
\end{figure}

\subsection{Calibration Performance on Driving Datasets}
\label{sec:exp_driving}
We conducted a quantitative evaluation of rotation and translation errors on the KITTI-360 and KITTI datasets.
As shown in Table~\ref{tab:calibration_errors}, GeoP-Calib achieved the lowest rotation error on 7 out of 10 sequences and the lowest translation error on 7 out of 10 sequences across KITTI-360 and KITTI.
In contrast, GST did not perform reliably on these driving scenarios, resulting in large errors and high variance. 
CLAIM, which leverages foundation models, attained relatively accurate translation in several sequences. 
However, its rotation accuracy was consistently inferior to the 3DGS-based approaches, limiting its overall calibration quality. 
Compared to prior 3DGS-based methods, GeoP-Calib yielded translation improvements over HiGS-Calib on every sequence, demonstrating that preserving the geometric integrity of the 3DGS proxy provides a consistently stronger calibration signal for metric alignment.
Specifically, GeoP-Calib improves over the second-best result by $0.019^\circ$ in rotation and $0.034$ m in translation on KITTI-360, and by $0.018^\circ$ in rotation and $0.011$ m in translation on KITTI.

\subsection{Ablation Study}
\label{sec:ablation}

Table~\ref{tab:ablation_component} reports the component-wise ablation of GeoP-Calib on KITTI-360. 
Starting from Base, adding DDA reduced the translation error from $0.077$\,m to $0.068$\,m, improving by $0.009$\,m, and slightly improved the rotation error from $0.122^\circ$ to $0.117^\circ$, improving by $0.005^\circ$. 
Adding GD alone also improved translation to $0.067$\,m, improving by $0.010$\,m, while rotation increased to $0.127^\circ$. However, the increase was marginal at $0.005^\circ$. 
When both DDA and GD are enabled, GeoP-Calib achieved the best overall performance, reducing translation to $0.063$\,m and rotation to $0.121^\circ$. 
Compared to Base, this yielded a $0.014$\,m improvement in translation and a $0.001^\circ$ improvement in rotation.

We additionally ablated the VSM used for dense depth anchoring. 
With the same DDA+GD setting, enabling VSM improved rotation from $0.124^\circ$ to $0.121^\circ$, improved by $0.003^\circ$, and improves translation from $0.066$\,m to $0.063$\,m, improving by $0.003$\,m. 
These results indicated that applying VSM for occlusion handling during dense depth anchoring provides a consistent benefit.

In addition, on our RTX~4070~Ti and Ryzen~3600 setup, adding DDA increased the runtime by about 50\,s compared to Base, resulting in an average per-sequence runtime of approximately 15.4\,minutes.


\begin{figure*}[t]
    \centering
    \begin{subfigure}[t]{0.32\textwidth}
        \centering
        \includegraphics[width=\linewidth]{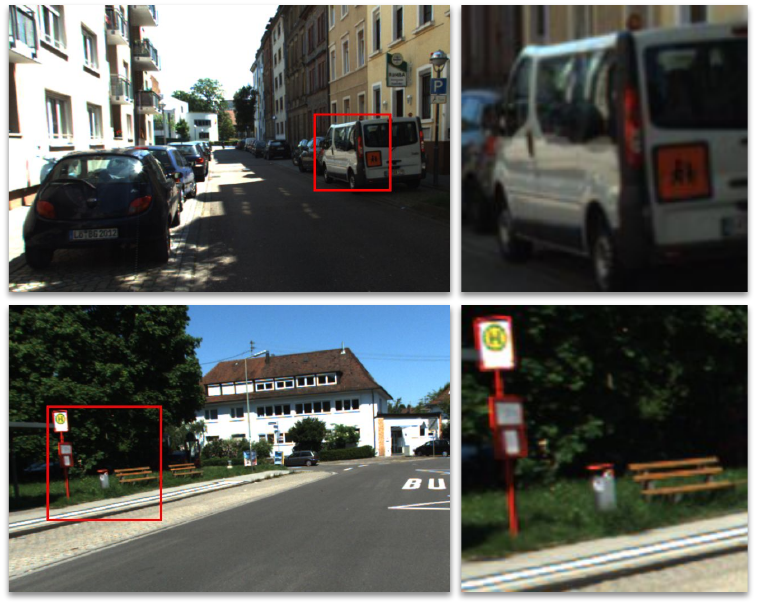} 
        \caption{}
        \label{fig:ablation_photometric_fidelity:gt}
    \end{subfigure}
    \begin{subfigure}[t]{0.32\textwidth}
        \centering
        \includegraphics[width=\linewidth]{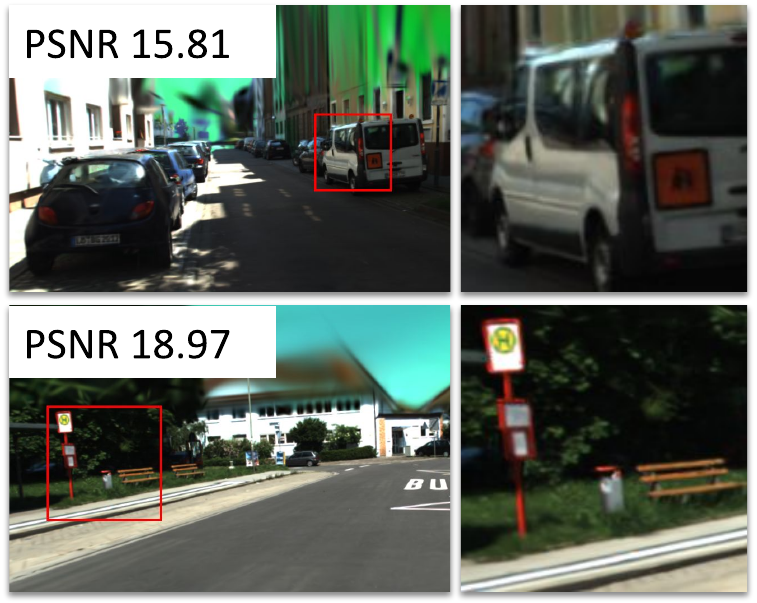} 
        \caption{}
        \label{fig:ablation_photometric_fidelity:base}
    \end{subfigure}
    \begin{subfigure}[t]{0.32\textwidth}
        \centering
        \includegraphics[width=\linewidth]{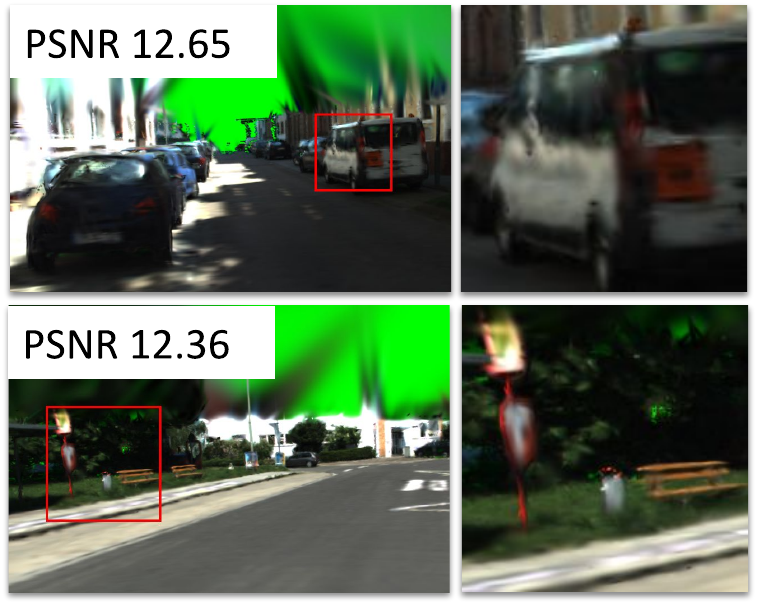}
        \caption{}
        \label{fig:ablation_photometric_fidelity:gp}
    \end{subfigure}
    \caption{
Qualitative comparison of rendered RGB images on KITTI-360: (\subref{fig:ablation_photometric_fidelity:gt}) GT, (\subref{fig:ablation_photometric_fidelity:base}) Base and (\subref{fig:ablation_photometric_fidelity:gp}) GeoP-Calib (Ours). PSNR is computed over the entire image to assess photometric consistency.
    }
    \label{fig:ablation_photometric_fidelity}
\end{figure*}

\subsection{Extended Analysis on Calibration Properties}
\label{sec:analysis}

\noindent\textbf{Robustness to Pose Initialization.}
We evaluated robustness to initialization on KITTI-360 Seq.~2 by injecting noise into the initial extrinsics. 
Specifically, we defined nine noise levels, where the rotation noise increased from $1^\circ$ to $9^\circ$, and the translation noise increased from $0.1$\,m to $0.9$\,m. 
For each noise level, we ran calibration from the perturbed initialization and reported the resulting rotation and translation errors.
Fig.~\ref{fig:noise-experiment} shows that GeoP-Calib converged robustly even under high initialization noise levels.

\noindent\textbf{Rendering Quality Does Not Imply Calibration Quality.}
We conducted this analysis on KITTI-360 Seq.~2 and Seq.~4. 
As shown in Fig.~\ref{fig:ablation_photometric_fidelity}, GeoP-Calib yielded lower PSNR than the Base pipeline, indicating reduced photometric consistency with the ground-truth RGB. 
Nevertheless, GeoP-Calib achieved higher LiDAR-Camera calibration accuracy on the same sequences. 
These results demonstrate that photometric fidelity is not a sufficient condition for accurate calibration. 
Rather, since extrinsic refinement is mediated through the Gaussian proxy, geometric faithfulness between the LiDAR structure and the 3DGS representation should be established as a prerequisite before improvements in photometric reconstruction can meaningfully translate into calibration gains.


\begin{figure*}[t]
    \centering
    \begin{subfigure}[t]{0.45\textwidth}
        \centering
        \includegraphics[width=\linewidth]{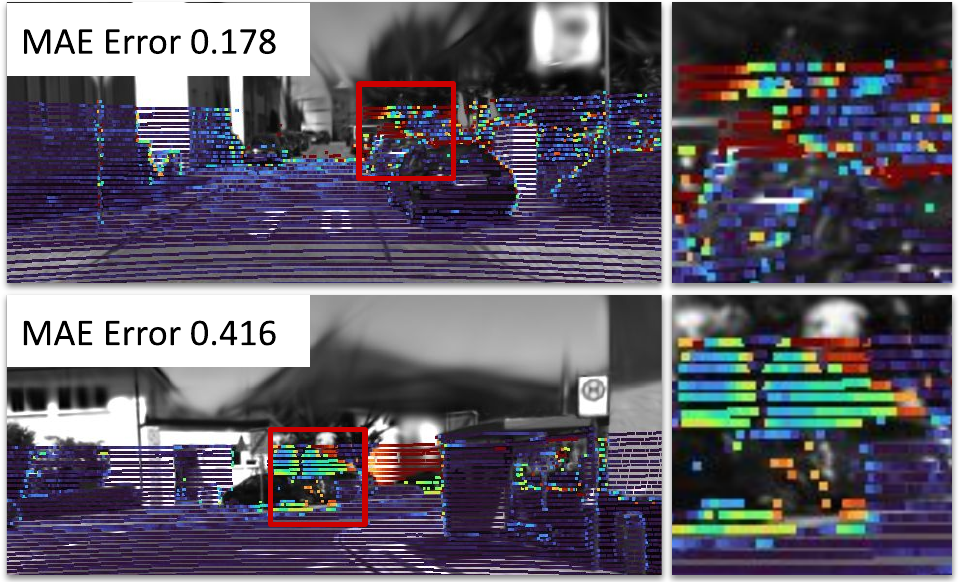} 
        \caption{}
        \label{fig:ablation_depth_accuracy:a}
    \end{subfigure}
    \begin{subfigure}[t]{0.45\textwidth}
        \centering
        \includegraphics[width=\linewidth]{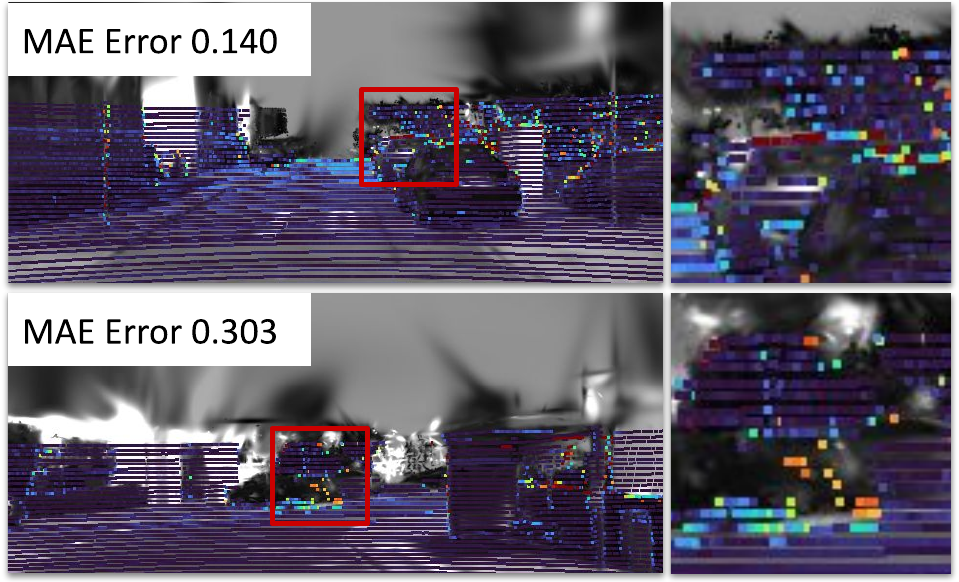} 
        \caption{}
        \label{fig:ablation_depth_accuracy:b}
    \end{subfigure}
    \begin{subfigure}[t]{0.053\textwidth}
        \centering
        \includegraphics[width=\linewidth]{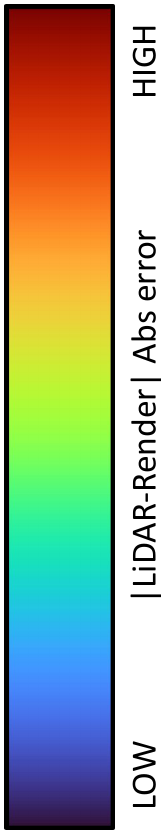} 
    \end{subfigure}
    \caption{
        Rendered depth error maps on KITTI-360, measured as the Mean Absolute Error (MAE) between rendered depth and LiDAR depth in the LiDAR-translation and camera-rotation setting: (\subref{fig:ablation_depth_accuracy:a}) Base and (\subref{fig:ablation_depth_accuracy:b}) GeoP-Calib (Ours).
        Colors indicate error magnitude, where blue denotes lower error and red denotes higher error.
    }
    \label{fig:ablation_depth_accuracy}
\end{figure*}

\noindent\textbf{Analysis for Geometric Accuracy.}
We evaluated how well GeoP-Calib preserves metric geometry compared to Base using the depth MAE between rendered depth and LiDAR depth on KITTI-360 after convergence. 
GeoP-Calib achieves a lower average MAE of $0.244$ than Base at $0.294$, corresponding to an absolute reduction of $0.05$ or approximately $17\%$. 
As shown in Fig.~\ref{fig:ablation_depth_accuracy}, Base exhibited localized regions where the depth error sharply increased, which is consistent with \textit{Geometric Decay} that distorts the proxy geometry under photometric optimization, whereas GeoP-Calib yields a more spatially consistent error distribution.



\section{Conclusion}
\label{conclusion}
We present GeoP-Calib, a geometry-preserving framework for targetless LiDAR-Camera calibration with 3DGS. By combining Dense Depth Anchoring and Gradient Decoupling, GeoP-Calib mitigates texture-driven \textit{Geometric Decay} while preserving LiDAR-consistent proxy geometry. Experiments on KITTI and KITTI-360 showed consistent calibration improvements, particularly in translation.

\noindent\textbf{Limitations and Future Works.}\enskip
While translation accuracy improved notably, rotation gains remained modest. 
Gradient Decoupling effectively defends against \textit{Geometric Decay}, but may limit the extent to which the optimization leverages photometric texture cues. 
A promising future direction is to maximize photometric consistency while still preventing \textit{Geometric Decay}, enabling stronger use of texture information without sacrificing metric faithfulness.



%
%
\bibliographystyle{splncs04}
\bibliography{main}

@String(CVPR  = {IEEE Conf. Comput. Vis. Pattern Recog.})

@String(ECCV  = {Eur. Conf. Comput. Vis.})

@String(AAAI  = {AAAI})

@String(CVPR  = {CVPR})

@String(ECCV  = {ECCV})

@article{kerbl20233d,
  title={3d gaussian splatting for real-time radiance field rendering.},
  author={Kerbl, Bernhard and Kopanas, Georgios and Leimk{\"u}hler, Thomas and Drettakis, George and others},
  journal={ACM Trans. Graph.},
  volume={42},
  number={4},
  pages={139--1},
  year={2023}
}

@article{mildenhall2021nerf,
  title={Nerf: Representing scenes as neural radiance fields for view synthesis},
  author={Mildenhall, Ben and Srinivasan, Pratul P and Tancik, Matthew and Barron, Jonathan T and Ramamoorthi, Ravi and Ng, Ren},
  journal={Communications of the ACM},
  volume={65},
  number={1},
  pages={99--106},
  year={2021},
  publisher={ACM New York, NY, USA}
}

@inproceedings{huang20242d,
  title={2d gaussian splatting for geometrically accurate radiance fields},
  author={Huang, Binbin and Yu, Zehao and Chen, Anpei and Geiger, Andreas and Gao, Shenghua},
  booktitle={ACM SIGGRAPH 2024 conference papers},
  pages={1--11},
  year={2024}
}

@article{kim2024camera,
  title={Camera-LiDAR extrinsic calibration using constrained optimization with circle placement},
  author={Kim, Daeho and Shin, Seunghui and Hwang, Hyoseok},
  journal={IEEE Robotics and Automation Letters},
  volume={10},
  number={2},
  pages={883--890},
  year={2024},
  publisher={IEEE}
}

@inproceedings{zhang2004extrinsic,
  title={Extrinsic calibration of a camera and laser range finder (improves camera calibration)},
  author={Zhang, Qilong and Pless, Robert},
  booktitle={2004 IEEE/RSJ International Conference on Intelligent Robots and Systems (IROS)(IEEE Cat. No. 04CH37566)},
  volume={3},
  pages={2301--2306},
  year={2004},
  organization={IEEE}
}

@inproceedings{levinson2013automatic,
  title={Automatic online calibration of cameras and lasers.},
  author={Levinson, Jesse and Thrun, Sebastian},
  booktitle={Robotics: science and systems},
  volume={2},
  number={7},
  year={2013},
  organization={Berlin, Germany}
}

@article{iyer2018calibnet,
  title={CalibNet: self-supervised extrinsic calibration using 3D spatial transformer networks},
  author={Iyer, Ganesh and Murthy, J Krishna and Krishna, K Madhava and others},
  journal={arXiv preprint arXiv:1803.08181},
  year={2018}
}

@inproceedings{zhou2018automatic,
  title={Automatic extrinsic calibration of a camera and a 3d lidar using line and plane correspondences},
  author={Zhou, Lipu and Li, Zimo and Kaess, Michael},
  booktitle={2018 IEEE/RSJ International Conference on Intelligent Robots and Systems (IROS)},
  pages={5562--5569},
  year={2018},
  organization={IEEE}
}

@article{yan2022opencalib,
  title={Opencalib: A multi-sensor calibration toolbox for autonomous driving},
  author={Yan, Guohang and Liu, Zhuochun and Wang, Chengjie and Shi, Chunlei and Wei, Pengjin and Cai, Xinyu and Ma, Tao and Liu, Zhizheng and Zhong, Zebin and Liu, Yuqian and others},
  journal={Software Impacts},
  volume={14},
  pages={100393},
  year={2022},
  publisher={Elsevier}
}

@inproceedings{zhu2020online,
  title={Online camera-lidar calibration with sensor semantic information},
  author={Zhu, Yufeng and Li, Chenghui and Zhang, Yubo},
  booktitle={2020 IEEE International Conference on Robotics and Automation (ICRA)},
  pages={4970--4976},
  year={2020},
  organization={IEEE}
}

@inproceedings{pandey2012automatic,
  title={Automatic targetless extrinsic calibration of a 3d lidar and camera by maximizing mutual information},
  author={Pandey, Gaurav and McBride, James and Savarese, Silvio and Eustice, Ryan},
  booktitle={Proceedings of the AAAI conference on artificial intelligence},
  volume={26},
  number={1},
  pages={2053--2059},
  year={2012}
}

@article{yuan2021pixel,
  title={Pixel-level extrinsic self calibration of high resolution lidar and camera in targetless environments},
  author={Yuan, Chongjian and Liu, Xiyuan and Hong, Xiaoping and Zhang, Fu},
  journal={IEEE Robotics and Automation Letters},
  volume={6},
  number={4},
  pages={7517--7524},
  year={2021},
  publisher={IEEE}
}

@inproceedings{zhang2021line,
  title={Line-based automatic extrinsic calibration of LiDAR and camera},
  author={Zhang, Xinyu and Zhu, Shifan and Guo, Shichun and Li, Jun and Liu, Huaping},
  booktitle={2021 IEEE International Conference on Robotics and Automation (ICRA)},
  pages={9347--9353},
  year={2021},
  organization={IEEE}
}

@inproceedings{sarlin20superglue,
  author    = {Paul-Edouard Sarlin and
               Daniel DeTone and
               Tomasz Malisiewicz and
               Andrew Rabinovich},
  title     = {{SuperGlue}: Learning Feature Matching with Graph Neural Networks},
  booktitle = {CVPR},
  year      = {2020}
}

@article{ou2023targetless,
  title={Targetless LiDAR-camera calibration via cross-modality structure consistency},
  author={Ou, Ni and Cai, Hanyu and Wang, Junzheng},
  journal={IEEE Transactions on Intelligent Vehicles},
  volume={9},
  number={1},
  pages={2636--2648},
  year={2023},
  publisher={IEEE}
}

@article{koide2023general,
  title={General, single-shot, target-less, and automatic lidar-camera extrinsic calibration toolbox},
  author={Koide, Kenji and Oishi, Shuji and Yokozuka, Masashi and Banno, Atsuhiko},
  journal={arXiv preprint arXiv:2302.05094},
  year={2023}
}

@inproceedings{zhang2025claim,
  title={CLAIM: Camera-LiDAR Alignment with Intensity and Monodepth},
  author={Zhang, Zhuo and Liu, Yonghui and Zhang, Meijie and Tan, Feiyang and Ding, Yikang},
  booktitle={2025 IEEE/RSJ International Conference on Intelligent Robots and Systems (IROS)},
  pages={17921--17926},
  year={2025},
  organization={IEEE}
}

@inproceedings{zhou2023inf,
  title={Inf: Implicit neural fusion for lidar and camera},
  author={Zhou, Shuyi and Xie, Shuxiang and Ishikawa, Ryoichi and Sakurada, Ken and Onishi, Masaki and Oishi, Takeshi},
  booktitle={2023 IEEE/RSJ International Conference on Intelligent Robots and Systems (IROS)},
  pages={10918--10925},
  year={2023},
  organization={IEEE}
}

@inproceedings{herau2024soac,
  title={Soac: Spatio-temporal overlap-aware multi-sensor calibration using neural radiance fields},
  author={Herau, Quentin and Piasco, Nathan and Bennehar, Moussab and Roldao, Luis and Tsishkou, Dzmitry and Migniot, Cyrille and Vasseur, Pascal and Demonceaux, C{\'e}dric},
  booktitle={Proceedings of the IEEE/CVF Conference on Computer Vision and Pattern Recognition},
  pages={15131--15140},
  year={2024}
}

@article{zhou2025robust,
  title={Robust LiDAR-Camera Calibration With 2D Gaussian Splatting},
  author={Zhou, Shuyi and Xie, Shuxiang and Ishikawa, Ryoichi and Oishi, Takeshi},
  journal={IEEE Robotics and Automation Letters},
  year={2025},
  publisher={IEEE}
}

@article{zhang2025higs,
  title={HiGS-Calib: A Hierarchical 3D Gaussian Splatting based Targetless Local-Consistent LiDAR-Camera Calibration Method},
  author={Zhang, Tianjun and Zhang, Lin and Wang, Hesheng},
  journal={IEEE Transactions on Circuits and Systems for Video Technology},
  year={2025},
  publisher={IEEE}
}

@inproceedings{herau20243dgs,
  title={3dgs-calib: 3d gaussian splatting for multimodal spatiotemporal calibration},
  author={Herau, Quentin and Bennehar, Moussab and Moreau, Arthur and Piasco, Nathan and Rold{\~a}o, Luis and Tsishkou, Dzmitry and Migniot, Cyrille and Vasseur, Pascal and Demonceaux, C{\'e}dric},
  booktitle={2024 IEEE/RSJ International Conference on Intelligent Robots and Systems (IROS)},
  pages={8315--8321},
  year={2024},
  organization={IEEE}
}

@misc{jung2026targetlesslidarcameracalibrationneural,
      title={Targetless LiDAR-Camera Calibration with Neural Gaussian Splatting}, 
      author={Haebeom Jung and Namtae Kim and Jungwoo Kim and Jaesik Park},
      year={2026},
      eprint={2504.04597},
      archivePrefix={arXiv},
      primaryClass={cs.CV}
}

@article{Liao2021ARXIV, 
   title   = {{KITTI}-360: A Novel Dataset and Benchmarks for Urban Scene Understanding in 2D and 3D}, 
   author  = {Yiyi Liao and Jun Xie and Andreas Geiger}, 
   journal = {arXiv preprint arXiv:2109.13410},
   year    = {2021}, 
}

@inproceedings{Geiger2012CVPR,
  author = {Andreas Geiger and Philip Lenz and Raquel Urtasun},
  title = {Are we ready for Autonomous Driving? The KITTI Vision Benchmark Suite},
  booktitle = {Conference on Computer Vision and Pattern Recognition (CVPR)},
  year = {2012}
}

@article{thrun2006stanley,
  title={Stanley: The robot that won the DARPA Grand Challenge},
  author={Thrun, Sebastian and Montemerlo, Mike and Dahlkamp, Hendrik and Stavens, David and Aron, Andrei and Diebel, James and Fong, Philip and Gale, John and Halpenny, Morgan and Hoffmann, Gabriel and others},
  journal={Journal of field Robotics},
  volume={23},
  number={9},
  pages={661--692},
  year={2006},
  publisher={Wiley Online Library}
}

@inproceedings{caesar2020nuscenes,
  title={nuscenes: A multimodal dataset for autonomous driving},
  author={Caesar, Holger and Bankiti, Varun and Lang, Alex H and Vora, Sourabh and Liong, Venice Erin and Xu, Qiang and Krishnan, Anush and Pan, Yu and Baldan, Giancarlo and Beijbom, Oscar},
  booktitle={Proceedings of the IEEE/CVF conference on computer vision and pattern recognition},
  pages={11621--11631},
  year={2020}
}

@inproceedings{lv2021lccnet,
  title={LCCNet: LiDAR and camera self-calibration using cost volume network},
  author={Lv, Xudong and Wang, Boya and Dou, Ziwen and Ye, Dong and Wang, Shuo},
  booktitle={Proceedings of the IEEE/CVF Conference on Computer Vision and Pattern Recognition},
  pages={2894--2901},
  year={2021}
}

@inproceedings{vora2020pointpainting,
  title={Pointpainting: Sequential fusion for 3d object detection},
  author={Vora, Sourabh and Lang, Alex H and Helou, Bassam and Beijbom, Oscar},
  booktitle={Proceedings of the IEEE/CVF conference on computer vision and pattern recognition},
  pages={4604--4612},
  year={2020}
}

@inproceedings{qiu2019deeplidar,
  title={Deeplidar: Deep surface normal guided depth prediction for outdoor scene from sparse lidar data and single color image},
  author={Qiu, Jiaxiong and Cui, Zhaopeng and Zhang, Yinda and Zhang, Xingdi and Liu, Shuaicheng and Zeng, Bing and Pollefeys, Marc},
  booktitle={Proceedings of the IEEE/CVF conference on computer vision and pattern recognition},
  pages={3313--3322},
  year={2019}
}

@inproceedings{yang2024unical,
  title={Unical: Unified neural sensor calibration},
  author={Yang, Ze and Chen, George and Zhang, Haowei and Ta, Kevin and B{\^a}rsan, Ioan Andrei and Murphy, Daniel and Manivasagam, Sivabalan and Urtasun, Raquel},
  booktitle={European Conference on Computer Vision (ECCV)},
  pages={327--345},
  year={2024},
  organization={Springer}
}

@inproceedings{song2024graphbev,
  title={Graphbev: Towards robust bev feature alignment for multi-modal 3d object detection},
  author={Song, Ziying and Yang, Lei and Xu, Shaoqing and Liu, Lin and Xu, Dongyang and Jia, Caiyan and Jia, Feiyang and Wang, Li},
  booktitle={European Conference on Computer Vision (ECCV)},
  pages={347--366},
  year={2024},
  organization={Springer}
}

@article{wang2023multi,
  title={Multi-modal 3d object detection in autonomous driving: a survey},
  author={Wang, Yingjie and Mao, Qiuyu and Zhu, Hanqi and Deng, Jiajun and Zhang, Yu and Ji, Jianmin and Li, Houqiang and Zhang, Yanyong},
  journal={International Journal of Computer Vision},
  volume={131},
  number={8},
  pages={2122--2152},
  year={2023},
  publisher={Springer}
}

@article{guan2025lidar,
  title={LiDAR-camera Cooperative Semantic Segmentation},
  author={Guan, He and Song, Chunfeng and Zhang, Zhaoxiang},
  journal={Machine Intelligence Research},
  volume={22},
  number={5},
  pages={956--968},
  year={2025},
  publisher={Springer}
}

@article{beltran2022automatic,
  title={Automatic extrinsic calibration method for LiDAR and camera sensor setups},
  author={Beltr{\'a}n, Jorge and Guindel, Carlos and De La Escalera, Arturo and Garc{\'\i}a, Fernando},
  journal={IEEE Transactions on Intelligent Transportation Systems},
  volume={23},
  number={10},
  pages={17677--17689},
  year={2022},
  publisher={IEEE}
}

@article{fan2025lidar,
  title={LiDAR, IMU, and camera fusion for simultaneous localization and mapping: A systematic review},
  author={Fan, Zheng and Zhang, Lele and Wang, Xueyi and Shen, Yilan and Deng, Fang},
  journal={Artificial Intelligence Review},
  volume={58},
  number={6},
  pages={174},
  year={2025},
  publisher={Springer}
}

@inproceedings{irie2016target,
  title={Target-less camera-LiDAR extrinsic calibration using a bagged dependence estimator},
  author={Irie, Kiyoshi and Sugiyama, Masashi and Tomono, Masahiro},
  booktitle={2016 IEEE International Conference on Automation Science and Engineering (CASE)},
  pages={1340--1347},
  year={2016},
  organization={IEEE}
}

@article{kang2020automatic,
  title={Automatic targetless camera--lidar calibration by aligning edge with gaussian mixture model},
  author={Kang, Jaehyeon and Doh, Nakju L},
  journal={Journal of Field Robotics},
  volume={37},
  number={1},
  pages={158--179},
  year={2020},
  publisher={Wiley Online Library}
}

@article{civera2008inverse,
  title={Inverse depth parametrization for monocular SLAM},
  author={Civera, Javier and Davison, Andrew J and Montiel, JM Martinez},
  journal={IEEE transactions on robotics},
  volume={24},
  number={5},
  pages={932--945},
  year={2008},
  publisher={IEEE}
}
\end{document}